\documentclass[sigconf,screen,nonacm]{acmart}
\usepackage[utf8]{inputenc}
\usepackage{xspace}
\usepackage{nicefrac}
\usepackage{soul}
\usepackage{xcolor}
\usepackage[capitalise]{cleveref}
\usepackage{amsmath}
\usepackage{amsfonts}
\usepackage{bm}
\usepackage{siunitx}
\usepackage{lipsum}
\usepackage{enumitem}
\usepackage[labelformat=simple]{subcaption}
\usepackage{dblfloatfix}      % table at bottom
\usepackage{tabu}
\usepackage{tabulary}
\usepackage{multirow}
\usepackage{multicol}
\usepackage{booktabs}
\usepackage{adjustbox}
\usepackage[flushleft]{threeparttable}
\usepackage{algorithm}
\usepackage{algpseudocodex}

\crefname{figure}{Figure}{Figures}  
\graphicspath{{figs/}}
\DeclareGraphicsExtensions{.pdf,.eps,.png,.jpg,.jpeg}

\newtheorem{theorem}{Theorem}
\newtheorem{property}[theorem]{Property}
\crefname{property}{Property}{Properties}  
\DeclareMathOperator{\softmax}{softmax}

\newcommand{\RR}[1]{\mathbb{R}^{#1}}

\newcommand{\ds}[1]{\textls[-60]{\textsc{\MakeLowercase{#1}}}}
\newcommand{\tpm}[1]{{\small{\textpm#1}}}
\newcommand{\mname}{\textsc{DHIL-GT}}

\newcommand{\blue}[1]{\textcolor{blue}{#1}}
\NewDocumentCommand{\leo}{mg}{\IfNoValueTF{#2}
{\textcolor{teal}{#1}}
{\textcolor{teal}{$\blacktriangleright$}#1 \textcolor{teal}{$\triangleright$ #2$\blacktriangleleft$}}}

\AtBeginDocument{%
  }

\setcopyright{acmlicensed}
\copyrightyear{2018}
\acmYear{2018}
\acmDOI{XXXXXXX.XXXXXXX}

\acmConference[Conference acronym 'XX]{Make sure to enter the correct
  conference title from your rights confirmation emai}{June 03--05,
  2018}{Woodstock, NY}
\acmISBN{978-1-4503-XXXX-X/18/06}

\begin{document}

\title[\mname{}: Scalable Graph Transformer with Decoupled Hierarchy Labeling]
{\mname{}: Scalable Graph Transformer\\with Decoupled Hierarchy Labeling}


\author{Ningyi Liao$^{\ast}$}
\affiliation{%
  \institution{Nanyang Technological University}
  \country{Singapore}}
\email{liao0090@e.ntu.edu.sg}

\author{Zihao Yu$^{\ast}$}
\myauthornote{Both authors contributed equally to this research.}
\affiliation{%
  \institution{Nanyang Technological University}
  \country{Singapore}
}
\email{zihao.yu@ntu.edu.sg}

\author{Siqiang Luo}
\affiliation{%
  \institution{Nanyang Technological University}
  \country{Singapore}}
\email{siqiang.luo@ntu.edu.sg}

\renewcommand{\shortauthors}{Trovato et al.}

\begin{abstract}
Graph Transformer (GT) has recently emerged as a promising neural network architecture for learning graph-structured data. However, its global attention mechanism with quadratic complexity concerning the graph scale prevents wider application to large graphs. While current methods attempt to enhance GT scalability by altering model architecture or encoding hierarchical graph data, our analysis reveals that these models still suffer from the computational bottleneck related to graph-scale operations. 
In this work, we target the GT scalability issue and propose \mname{}, a scalable Graph Transformer that simplifies network learning by fully decoupling the graph computation to a separate stage in advance. \mname{} effectively retrieves hierarchical information by exploiting the graph labeling technique, as we show that the graph label hierarchy is more informative than plain adjacency by offering global connections while promoting locality, and is particularly suitable for handling complex graph patterns such as heterophily. We further design subgraph sampling and positional encoding schemes for precomputing model input on top of graph labels in an end-to-end manner. The training stage thus favorably removes graph-related computations, leading to ideal mini-batch capability and GPU utilization. Notably, the precomputation and training processes of \mname{} achieve complexities \textit{linear} to the number of graph edges and nodes, respectively. 
Extensive experiments demonstrate that \mname{} is efficient in terms of computational boost and mini-batch capability over existing scalable Graph Transformer designs on large-scale benchmarks, while achieving top-tier effectiveness on both homophilous and heterophilous graphs. 
\end{abstract}



\received{20 February 2007}
\received[revised]{12 March 2009}
\received[accepted]{5 June 2009}

\maketitle

% !TEX root = ../main.tex

\section{Introduction}
\label{sec:introduction}

Graph Transformers characterize a family of neural networks that introduce the powerful Transformer architecture \cite{Vaswani2017Attention} to the realm of graph data learning. These models have garnered increasing research interest due to their unique applications and competitive performance \cite{ying2021graphormer,wu2021graphtrans,hussain2022egt,zhu2023a}. Despite their achievements, vanilla GTs are highly limited to specific tasks because of the full-graph attention mechanism, which has computational complexity at least quadratic to the graph size, rendering it impractical for a single graph with more than thousands of nodes. Enhancing the scalability of GTs is thus a prominent task for enabling these models to handle a wider range of graph data on large scales.

To scale up Graph Transformers, existing studies explore various strategies to retrieve and utilize graph data efficiently. One representative approach is to simplify the model architecture with a specialized attention module \cite{wu2022nodeformer,rampavsek2022graphgps,wu2023difformer}. The graph topology is conserved by the message-passing mechanism, which recognizes edge connections without the need for quadratic computation on all-pair node interactions. However, these modifications introduce another computational bottleneck of iterative graph propagation, which typically has an overhead linear to the edge size and remains challenging for scalable model training.
An alternative line of works chooses to embed richer topological information as structured data through different graph processing techniques, such as adjacency-based spatial propagation \cite{chen2023c,kong2023goat}, polynomial spectral transformation \cite{ma2023polyformer,deng2024polynormer}, and hierarchical graph coarsening \cite{zhang2022ansgt,zhu2023hsgt}. Although these models offer a relatively scalable model training scheme, the graph-related operation still persists during their learning pipeline, leaving certain scalability and expressivity issues unsolved as detailed in our complexity analysis.

\begin{table*}[t]
% \captionsetup{font=normalsize}
\caption{Time and memory complexity of different types of Graph Transformer models with respect to precomputation and training stages. 
``FB'', ``NS'', and ``RS'' refer to full-batch, neighborhood sampling, and random sampling strategies, respectively. 
Training time complexity represent the forward-passing computational operations on respective node sets, while precomputation complexity indicates one-time processing such as positional encoding and token generation. RAM memory represents the primary data used as input, and GPU memory is for variable graph or batch representations during learning.  ``Hetero'' column marks whether the model evaluates data under heterophily in the original literature.
% We also append the largest graph size used by the original papers indicated by node size $n$ and edge size $m$. 
}
\label{tab:complexity}
\renewcommand{\arraystretch}{1.12}
\begin{adjustbox}{max width=\textwidth}
\centering
% \small
% \setlength{\tabcolsep}{1.8pt}
\begin{tabular}{ccc|cc|cc|c}
\toprule
    \textbf{Taxonomy} & \textbf{Batch} & \textbf{Model}
            & \textbf{Precompute Time} & \textbf{Train Time} 
            & \textbf{RAM Mem.} & \textbf{GPU Mem.} & \textbf{Hetero} \\
\hline
\multirow{2}{*}{Vanilla} & \multirow{2}{*}{FB} 
	% & GraphTrans~\cite{wu2021graphtrans}   & -- & $ O(Ln^2F)   $    & & & 0.3K  \\    
    & Graphormer~\cite{ying2021graphormer} & $ O(n^3) $ & $ O(Ln^2F) $ & $O(n^2)$ & $ O(Ln^2F) $ & N \\
    & & GRPE~\cite{park2022grpe} & $O(n^2)$ & $O(Ln^2F)$ & $O(n^2)$ & $O(Ln^2F)$ &  N \\
\midrule
\multirow{4}{*}{Kernel-based} & \multirow{4}{*}{NS} 
    & GraphGPS~\cite{rampavsek2022graphgps} & $O(n^3)$ & $O(LnF^2 + LmF)$ & $O(nF+n^2)$ & $O(Ln_bF+m)$ & Y  \\
    & & NodeFormer~\cite{wu2022nodeformer} & -- & $ O(LnF^2+LmF) $ & $O(nF+m)$ & $O(Ln_bF+m)$ & Y  \\
    & & DIFFormer~\cite{wu2023difformer}  & -- & $ O(LnF^2+LmF) $ & $O(nF+m)$ & $O(Ln_bF+m)$ & N  \\
    % & & SGFormer~\cite{wu2023sgformer}  & -- & $ O(LnF^2+LmF) $ & 0.1B \\
    & & PolyNormer~\cite{deng2024polynormer} & -- & $ O(LnF^2+LmF)$ & $O(nF+m)$ & $O(Ln_bF+m)$ & Y \\
\midrule
\multirow{6}{*}{Hierarchical} & \multirow{6}{*}{RS} 
    & NAGphormer~\cite{chen2023c} & $O(LmF_0)$ & $O(LnF^2)$ & $O(LnF)$ & $O(Ln_bF^2)$ & N \\
    & & PolyFormer~\cite{ma2023polyformer} & $O(LmF_0)$ & $ O(LnF^2)$ & $O(LnF)$ & $O(Ln_bF^2)$ & Y \\
    & & ANS-GT~\cite{zhang2022ansgt} & $O(ns^2+Lm)$ & $O(LnF^2+Lns^2F+)$ & $O(nF+ns^2)$ & $O(Ln_bF+n_bs^2)$ & Y \\
    & & GOAT~\cite{kong2023goat} & $O(nF)$ & $O(LnF^2+LmF) $ & $O(nF+m)$ & $O(Ln_b^2+Ln_bF+m)$ & Y \\
    & & HSGT~\cite{zhu2023hsgt} & $O(n+Lm)$ & $O(LnF^2+LmF) $ & $O(nF+Lm)$ & $O(Ln_b^2+Ln_bF+Lm)$ & N \\
    & & \textbf{\mname{} (ours)} & $O(ns^3+ms)$ & $O(LnF^2+Lns^2F) $ & $O(nF+ns^2)$ & $O(Ln_bF+n_bs^2)$ & Y \\
\bottomrule
\end{tabular}
\end{adjustbox}
% \vspace{-1.8ex}
\end{table*}

In this work, we propose \mname{}, a scalable Graph Transformer with \ul{\textsc{D}}ecoupled \ul{\textsc{Hi}}erarchy \ul{\textsc{L}}abeling.
By constructing a hierarchical graph label set consisting of node pair connections and distances, we showcase that all graph information necessary for GT learning can be fully decoupled and produced by an end-to-end pipeline before training. The precomputation procedure can be finished within a linear $O(m)$ bound, while the iterative learning step is as simple as training normal Transformers with $O(n)$ complexity, where $m$ and $n$ are the numbers of graph edges and nodes, respectively. The two stages achieve theoretical complexities on par with respective state-of-the-art GTs, as well as a substantial boost in practice thanks to empirical acceleration strategies.

Our \mname{} is based on the 2-hop labeling technique, which has been extensively studied with scalable algorithms for querying the shortest path distance (SPD) between nodes \cite{akiba2013,yano2013fast,akiba2015efficient}. By investigating its properties, we show that the graph labels construct a hierarchical representation of the graph topology, favorably containing both local and global graph connections. We design a novel subgraph token generation process to utilize the labels as informative input for GT. The data hierarchy benefits GT expressivity in modeling node-pair interactions beyond graph edges, which is superior in capturing graph knowledge under both homophily and heterophily compared to current locality-based GTs. 
In addition, the built graph labels also offer a simple and fast approach to query pair-wise distance as positional encoding. 
% By exploiting the built labels, positional encoding and feature augmentation can be easily acquired as well. 
To this end, graph information is decently embedded into the precomputed data from different perspectives, and all computations concerning graph labels can be performed in a one-time manner with efficient implementation. 
The GT training stage only learns from the structured input data, which enjoys ideal scalability, including a simple mini-batch scheme and memory overhead free from the graph size.

We summarize the contributions of this work as follows:
\begin{itemize}[wide,labelwidth=!,labelindent=0pt,itemsep=0pt,topsep=0mm]
    \item We propose \mname{} as a scalable Graph Transformer with decoupled graph computation and simple model training independent of graph operations. Both precomputation and learning stages achieve complexities only \textit{linear} to the graph scale. 
    \item We introduce an end-to-end precomputation pipeline for \mname{} based on graph labeling, efficiently embedding graph information with informative hierarchy. Dedicated token generation, positional encoding, and model architectures are designed for representing hierarchical data at multiple levels.
    \item We conduct comprehensive experiments to evaluate the effectiveness and efficiency of \mname{} against current Graph Transformers across large-scale homophilous and heterophilous graphs. \mname{} achieves top-tier accuracy and demonstrates competitive scalability regarding time and memory overhead.
\end{itemize}

\section{Related works}
\label{sec:related}

\noindentparagraph{Vanilla Graph Transformers.} 
Early GTs ~\cite{dwivedi2020generalization,ying2021graphormer,wu2021graphtrans} are mainly proposed for graph-level learning tasks, typically involving small-scale graphs of less than a thousand nodes. Following the vanilla design, a wide range of positional encoding schemes have been invoked to the self-attention module to encode graph topology information, including graph proximity~\cite{ying2021graphormer,zhang2022ansgt, chen2021litegt}, Laplacian eigenvectors~\cite{dwivedi2020generalization,kreuzer2021rethinking,hussain2022egt}, and shortest path distance~\cite{park2022grpe,chen2022sat,zhao2023deepgraph}.

As listed in \cref{tab:complexity}, the critical scalability bottleneck of these models lies in the straight-forward attention mechanism calculating all node-pair interactions in the graph, resulting $O(n^2)$ complexity of both training time and memory. If there is positional encoding, additional preprocessing is also demanded with $O(n^2)$ or even higher overhead. 
A naive solution is to randomly sample a subset of nodes and adopt mini-batch learning. However, it largely overlooks graph information and results in suboptimal performance. 

\noindentparagraph{Kernel-based Graph Transformers.} 
Kernelization is the method for modeling node-pair relations and replacing the vanilla self-attention scheme. For instance, NodeFormer~\cite{wu2022nodeformer} employs a kernel function based on random features, and GraphGPS~\cite{rampavsek2022graphgps} opts to incorporate topological representation. More expressive kernels are also developed, invoking depictions such as graph diffusion~\cite{wu2023difformer} and node permutation~\cite{deng2024polynormer}. 

Although kernelized GTs prevent the quadratic complexity, the nature of the kernel indicates that graph data needs to be iteratively accessed during learning, which is represented by the $O(LmF)$ learning overhead in \cref{tab:complexity}. When the graph scale is large, this term becomes dominant since the edge size $m$ is significantly larger than the node size $n$. Hence, we argue that such a design is not sufficiently scalable. 
Another under-explored issue is the expressiveness of the neighborhood sampling (NS) strategy for forming node batches in kernel-based GTs. Similar to convolutional GNNs, NS is known to be subject to performance loss on complex graph signals due to its inductive bias on graph homophily \cite{Breuer2020,zheng2022b}.

\noindentparagraph{Hierarchical Graph Transformers.} 
Recent advances reveal that it is possible to remove the full-graph attention and exploit the power of GTs to learn the latent node relations during learning. This is achieved by providing sufficient hierarchical context as input data with node-level identity. The key of this approach is crafting an effective embedding scheme to comply with GT expressivity. To realize this, NAGphormer~\cite{chen2023c}, PolyFormer~\cite{ma2023polyformer}, and GOAT~\cite{kong2023goat} look into representative features using adjacency propagation, spectral graph transformation, and feature space projection, respectively. ANS-GT~\cite{zhang2022ansgt} builds graph hierarchy by adaptive graph sampling concerning subgraphs of size $s$, while HSGT~\cite{zhu2023hsgt} leverages graph coarsening algorithms. 

Hierarchical GTs are applicable to mini-batching with random sampling (RS) as long as their graph embeddings are permutation invariant. 
Furthermore, since graph processing is independent of GT attention, it can be adequately improved with better algorithmic scalability. In most scenarios, the graph can be processed in $O(m)$ complexity in precomputation as shown in \cref{tab:complexity}. Nonetheless, we note that hierarchical models, except for NAGphormer and PolyFormer, still involve graph-level operations during training, which hinders GPU utilization and causes additional overhead. 

\noindentparagraph{Scalable Convolutional GNNs.} 
The scalability issue has also been extensively examined for Graph Neural Networks (GNNs) exploiting graph convolutions \cite{kipf2016semi,velivckovic2017graph}. Similar to hierarchical GTs, decoupled models propose to separate the graph computation from iterative convolution and employ dedicated acceleration, exhibiting excellent scalability on some of the largest datasets with linear or even sub-linear complexity \cite{Klicpera2019,wu19sim,Chen2020a,liao2022}. It is also demonstrated that such strategy is capable of handling heterophily \cite{li2021f,wang2022b,liao2023a}. Graph simplification techniques including graph sampling \cite{chen_fastgcn_2018,chiang2019cluster,zou_layer-dependent_2019,feng2022} and coarsening \cite{deng_graphzoom_2020,huang_scaling_2021,cai_graph_2021} approaches are also explored for reducing the graph scale at different hierarchy levels. 
Although the high-level idea of scaling up convolutional GNNs is helpful towards scalable GTs, Transformer-based models are unique in respect to their graph data utilization and architectural bottlenecks, and hence require specific designs for addressing their scalability issues. 

\section{Preliminaries}
\label{sec:preliminaries}

\noindentparagraph{Graph Labeling.} 
Consider a connected graph $\mathcal{G} = \langle \mathcal{V}, \mathcal{E}\rangle$ with $n = |\mathcal{V}|$ nodes and $m = |\mathcal{E}|$ edges. The node attribute matrix is $\bm{X} \in \RR{n \times F_0}$, where $F_0$ is the dimension of input attributes. 
The neighborhood of a node $v$ is $\mathcal{N}(v) = \{u | (v, u) \in \mathcal{E}\}$, and its degree $d(v) = |\mathcal{N}(v)|$. $\mathcal{P}(u,v)$ denotes a path from node $u$ to $v$, and the shortest distance $b(u,v)$ is achieved by the path with least nodes. 

% Each node $v\in\mathcal{V}$ is assigned with a label containing a node set $\mathcal{L}(v)$ and corresponding shortest distances $b(u,v), u\in\mathcal{L}(v)$.
The graph labeling process assigns a label $\mathcal{L}(v)$ to each node $v\in\mathcal{V}$, which is a set of pairs $(u, \delta)$ containing certain nodes $u$ and corresponding shortest distances $\delta = b(u,v)$ between the node pairs. The graph labels compose a 2-hop cover \cite{Cohen2003Reachability} of $\mathcal{G}$ if for an arbitrary node pair $u,v\in \mathcal{V}$, there exists $w\in\mathcal{L}(u), w\in\mathcal{L}(v)$ and $b(u,v) = b(u,w) + b(w,v)$. 
Given an order of all nodes in $\mathcal{V}$, we denote each node by a unique index $1, \cdots, n$ and use $u < v$ to indicate that node $u$ precedes node $v$ in the sequence. 

\noindentparagraph{Transformer Architecture.} 
A Transformer layer~\cite{Vaswani2017Attention} projects the input representation matrix $\bm{H}\in \RR{n\times F}$ into three subspaces:
\begin{equation}
    \bm{Q} = \bm{H}\bm{W}_Q,\quad 
    \bm{K} = \bm{H}\bm{W}_K,\quad 
    \bm{V} = \bm{H}\bm{W}_V,
\label{eq:attn}
\end{equation}
where $\bm{W}_Q\in\RR{F\times d_K}$, $\bm{W}_K\in\RR{F\times d_K}$, $\bm{W}_V\in\RR{F\times d_V}$ are the projection matrices. 
For a multi-head self-attention module with $N_H$ heads, each attention head possesses its own representations $\bm{Q}_i, \bm{K}_i, \bm{V}_i, i=1,\cdots,N_H$, and the output $\Tilde{\bm{H}}$ across all heads is calculated as:
\begin{equation}
    \Tilde{\bm{H}}_i = \softmax\left( \frac{\bm{Q}_i\bm{K}_i^\top}{\sqrt{d_K}} \right) \bm{V}_i,\quad
    \Tilde{\bm{H}} = (\Tilde{\bm{H}}_1 \| \cdots \| \Tilde{\bm{H}}_{N_H}) \,\bm{W}_O,
\label{eq:gt}
\end{equation}
where $\cdot\|\cdot$ denotes the matrix concatenation operation. In this paper, we set the projection dimension $d_K = d_V = F / N_H$. 
It can be observed that \cref{eq:attn,eq:gt} for representations of $n$ nodes lead to $O(n^2F)$ time and memory overhead. When it only applies to a batch of $n_b$ nodes, the complexity is drastically reduced to $O(n_b^2F)$. 

% !TEX root = ../main.tex

\begin{figure}[tp]
\captionsetup[subfigure]{skip=4pt}
\centering
    \includegraphics[width=\columnwidth]{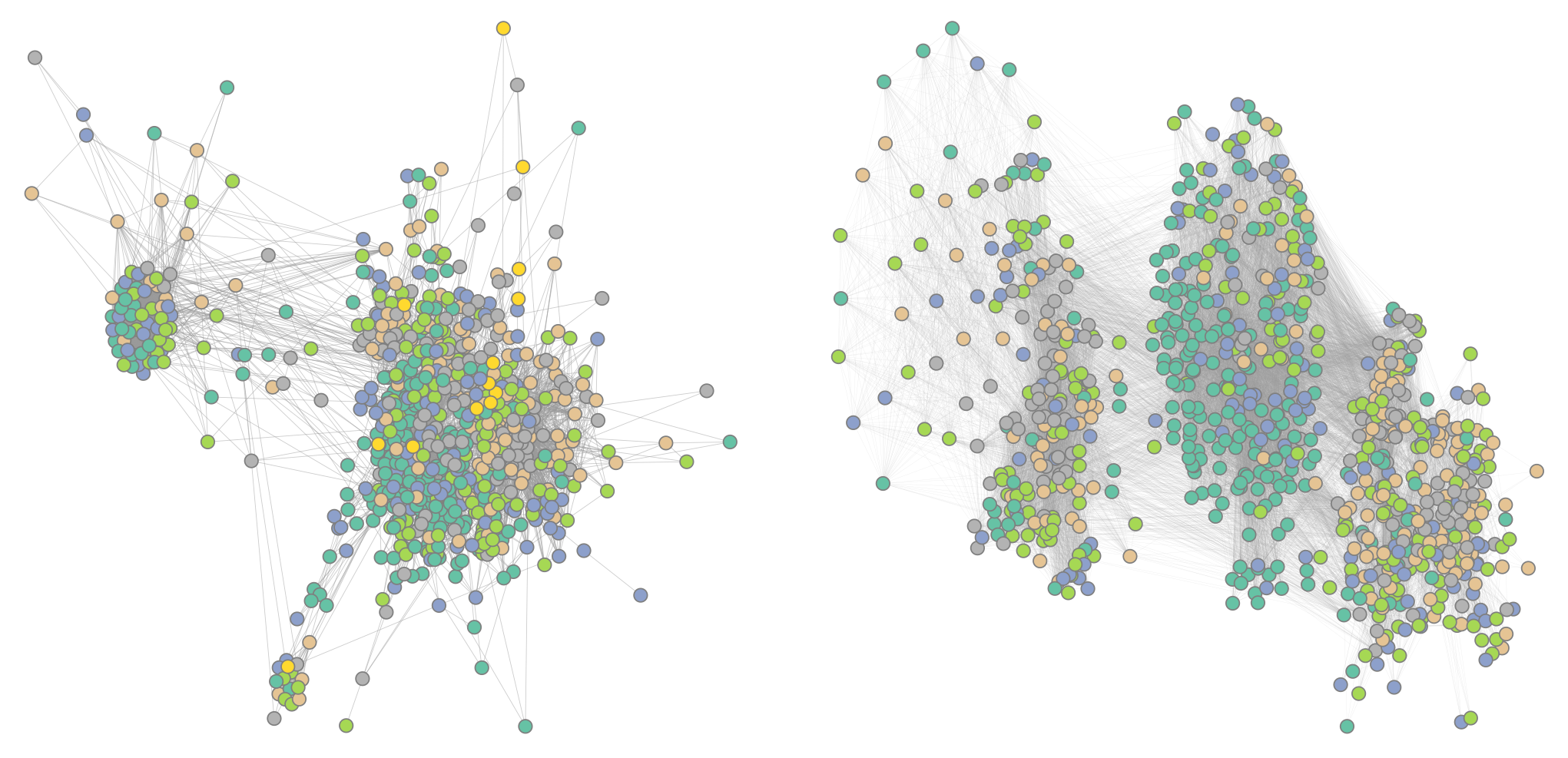}\vspace{-12pt}\\
    % \hfill
    \subcaptionbox{\ds{chameleon} Raw Graph\label{fig:comm_chameleon_filtered}}%
    [0.49\columnwidth]{}
\hfil
    \subcaptionbox{\ds{chameleon} Label Graph\label{fig:colb_chameleon_filtered}}%
    [0.49\columnwidth]{}
\caption{Visualization of the hierarchy of (a) the original graph $\mathcal{G}$ and (b) the label graph $\Hat{\mathcal{G}}$ on the heterophilous dataset \ds{chameleon}. Color of each node denotes its class.}
\label{fig:comm}
\vspace{-1.0em}
\end{figure}

% ====================
\section{Hierarchy of Label Graph}
\label{sec:pll}
Our \mname{} aims to retrieve graph hierarchy information from graph labels consisting of node pair connections and distances. In this section, we first introduce the pruned landmark labeling algorithm to efficiently compute graph labels as a 2-hop cover. Then, we analyze that the labeling process favorably builds a graph hierarchy with several useful properties for representing implicit graph information beyond adjacency. 

% ==========
\subsection{Pruned Landmark Labeling}
Based on the concept of graph labeling in \cref{sec:preliminaries}, a straight-forward approach to build the graph labels is to traverse the whole graph for each node successively. This is, however, prohibitive due to the repetitive graph traversal procedure. Hence, we employ the Pruned Landmark Labeling (PLL) algorithm \cite{akiba2013}, which constructs labels with a more efficient search space.

The PLL algorithm is presented in \cref{alg:label}. It performs a pruned BFS for each node indexed $1$ to $n$ following the given order. The algorithm is agnostic to the specific search order. In this work, we follow \cite{akiba2013} to adopt the descending order of node degrees for its satisfying performance while leaving other schemes for future exploration.

The PLL is more efficient than full-graph traversal as it prevents the visit to nodes $u$ that have been accessed and labeled with a shorter distance $b(u,v)$ to the current source node. Intuitively, during the early rounds of pruned BFS starting from nodes $v$ with smaller indices, the traversal is less pruned and can reach a large portion of the graph. These nodes are regarded as landmarks with higher significance and are able to appear in a large number of node labels $\mathcal{L}(u)$ where $u > v$ along with the distance information. On the contrary, for latter nodes with higher indices, the pruned traversal constrains the visit to the local neighborhood.

Thus, we reckon that the PLL process naturally builds a hierarchy embedded in the node labels. An exemplary illustration of a real-world graph is displayed in \cref{fig:comm}. The original \ds{chameleon} graph is heterophilous, i.e., connected nodes frequently belong to distinct classes. In \cref{fig:comm_chameleon_filtered}, different classes are mixed in graph clusters, which pose a challenge for GTs to perform classification based on edge connections. In contrast, nodes in the graph marked by graph labels in \cref{fig:colb_chameleon_filtered} clearly form multiple densely connected clusters, exhibiting a distinct hierarchy. Certain classes can be intuitively identified from the hierarchy, which empirically demonstrates the effectiveness of our utilization of graph labeling.

{%\small 
\begin{algorithm}[bp]
\algrenewcommand{\alglinenumber}[1]{\scriptsize\bfseries#1}
% \fontsize{9pt}{12pt}\selectfont
\renewcommand{\algorithmicrequire}{\textbf{Input:}}
\renewcommand{\algorithmicensure}{\textbf{Output:}}
\caption{Pruned Landmark Labeling \cite{akiba2013}}
\label{alg:label}
\begin{algorithmic}[1]
\Require Graph $\mathcal{G} = \langle \mathcal{V}, \mathcal{E}\rangle$
\Ensure Labels for all nodes $\mathcal{L}$
\State Sort $\mathcal{V}$ based on degree $d(v)$
\State $\mathcal{L}(v) \gets \varnothing$ for all $v\in\mathcal{V}$
\For{$v=1$ to $n$}
    \State Queue $\mathcal{Q} \gets \{(v, 0)\}$
    \While{$\mathcal{Q} \neq \varnothing$}
        \State Pop the first element $(u, \delta)$ from $\mathcal{Q}$
        \State $b(u, v) \gets \min\{b(u,w)+b(w,v) \,|\, w\in\mathcal{L}(u) \cap \mathcal{L}(v)\}$
        \If{$\delta < b(u, v)$}
            \State $\mathcal{L}(u) \gets \mathcal{L}(u) \cup {(v, \delta)}$
            \ForAll{$w\in\mathcal{N}(u)$ such that $w > v$}
                \State Push $(w, \delta+1)$ to the end of $\mathcal{Q}$
            \EndFor
        \EndIf
    \EndWhile
\EndFor\vspace{2pt}
\State \Return {$\{\mathcal{L}(v) \,|\, v\in\mathcal{V}\}$ as $\mathcal{L}$}
\end{algorithmic}
\end{algorithm}
}

% ==========
\subsection{Label Graph Properties}
\label{ssec:pll_property}
Then, we formulate the hierarchy in graph labels by defining a generated graph, namely the \textit{label graph}, as $\Hat{\mathcal{G}} = \langle \mathcal{V}, \Hat{\mathcal{E}} \rangle$, which is with directed and weighted edges. Its edge set depicts the elements in node labels computed by graph labeling, that an edge $(u,v) \in \Hat{\mathcal{E}}$ if and only if $(v,\delta)\in\mathcal{L}(u)$, and the edge weight is exactly the distance in graph labels $\delta = b(u,v)$. The in- and out-neighborhoods based on edge directions are $\mathcal{N}_{in}(v) = \{u | (u, v) \in \Hat{\mathcal{E}}\}$ and $\mathcal{N}_{out}(v) = \{u | (v, u) \in \Hat{\mathcal{E}}\}$, respectively. 

\begin{figure}[tp]
    \centering
    \includegraphics[width=\columnwidth]{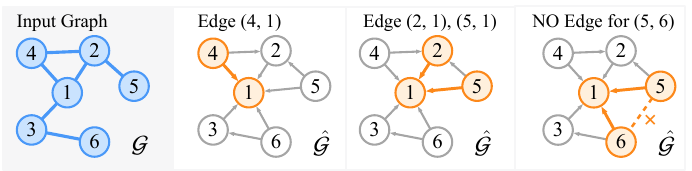}\vspace{-12pt}\\
    % \hfill
    \subcaptionbox{Raw Graph\label{fig:ex0}}%
    [0.24\columnwidth]{}
    \subcaptionbox{\cref{thm:neighbor}\label{fig:ex1}}%
    [0.24\columnwidth]{}
    \subcaptionbox{\cref{thm:path}\label{fig:ex2}}%
    [0.24\columnwidth]{}
    \subcaptionbox{\cref{thm:subgraph}\label{fig:ex3}}%
    [0.24\columnwidth]{}
    \caption{Examples of three properties of the label graph $\Hat{\mathcal{G}}$ corresponding to the original graph $\mathcal{G}$. Number inside each node denotes its index in descending order of node degrees. }
    \label{fig:ex}
% \vspace{-1.0em}
\end{figure}

We then elaborate the following three hierarchical properties of the label graph generated by \cref{alg:label}. Corresponding running examples are given in \cref{fig:ex}. For simplicity, we assume that the original graph $\mathcal{G}$ is undirected, while properties for a directed $\mathcal{G}$ can be acquired by separately considering two label sets $\mathcal{L}_{in}$ and $\mathcal{L}_{out}$ for in- and out-edges in $\mathcal{E}$. 

\begin{property}
\label{thm:neighbor}
    For an edge $(u,v)\in\mathcal{E}$, there is $(v,u) \in \Hat{\mathcal{E}}$ when $u<v$, and $(u,v) \in \Hat{\mathcal{E}}$ when $u>v$. 
\end{property}
Referring to \cref{alg:label}, when the current node is $v$ and $v < u$, $\delta = 1$ holds since $u$ is the direct neighbor of $v$. Hence, $(v, 1)$ is added to label $\mathcal{L}(u)$ at this round, which is equivalent to adding edge $(u,v)$ to $\Hat{\mathcal{E}}$. Similarly, $(v,u) \in \Hat{\mathcal{E}}$ holds when $v > u$. For example, the edge $(1, 4)$ in \cref{fig:ex0} is represented by the directed edge $(4,1)$ in \cref{fig:ex1}. 
\cref{thm:neighbor} implies that $\mathcal{N}(v) \subset \mathcal{N}_{in}(v) \cup \mathcal{N}_{out}(v)$, i.e., the neighborhood of the original graph is also included in the label graph, and is further separated into two sets according to the relative order of neighboring nodes. 

\begin{property}
\label{thm:path}
    For a shortest path $\mathcal{P}(u,v)$ in $\mathcal{G}$, there is $(w,v) \in \Hat{\mathcal{E}}$ for each $w \in \mathcal{P}(u,v)$ satisfying $w > v$. 
\end{property}
\cite{akiba2013} proves that there is $v\in\mathcal{L}(w)$ for $w \in \mathcal{P}(u,v)$ and $w > v$. Therefore, considering shortest paths starting with node $v$ of a small index, i.e., $v$ being a ``landmark'' node, then succeeding nodes $w > v$ in the path are connected to $v$ in $\Hat{\mathcal{G}}$. In \cref{fig:ex0}, the shortest path between $(1,5)$ passing node $2$ results in edges $(2,1)$ and $(5,1)$ in \cref{fig:ex2}, since nodes $2$ and $5$ are in the path and their indices are larger than node $1$. 
When the order is determined by node degree, high-degree nodes appear in shortest paths more frequently, and consequently link to a majority of nodes, including those long-tailed low-degree nodes in $\Hat{\mathcal{G}}$. 

\begin{property}
\label{thm:subgraph}
    For a shortest path $\mathcal{P}(u,v)$ in $\mathcal{G}$, if there is $w \in\mathcal{P}(u,v)$ and $w < v$, then $(u,v) \notin \Hat{\mathcal{E}}$. 
\end{property}
According to the property of shortest path, there is $b(u,v) = b(u,w) + b(w,v)$. Hence, the condition of line 8 in \cref{alg:label} is not met at the $v$-th round when visiting $w$. In other words, the traversal from $v$ is pruned at the preceding node $w$. By this means, the in-neighborhood $\mathcal{N}_{in}(v)$ is limited in the local subgraph with shortest paths ending at landmarks. 
As shown in \cref{fig:ex3}, the shortest path between $(5,6)$ passes node $1$, indicating that $(5,6)$ are not directly connected since their distance can be acquired by edges $(5,1)$ and $(6,1)$. As a consequence, the neighborhood of node $5$ in $\Hat{\mathcal{G}}$ is constrained by nodes $1$ and $2$, preventing connections to more distant nodes such as $3$ or $6$. 

Summarizing \cref{thm:neighbor,thm:path,thm:subgraph}, the label graph preserves neighboring connections of the original graph, while establishing more connections to a minority set of global nodes as landmark. The hierarchy is built so that long-tailed nodes with high indices are usually located in local substructures separated by landmarks. 
Noticeably, since the label graph is deterministic, it can be computed by \cref{alg:label} in an individual stage in one time and used throughout graph learning iterations. 

\begin{figure*}[!t]
% \vspace{-6mm}
% \captionsetup{skip=8pt}
\centering
\includegraphics[width=0.9\textwidth]{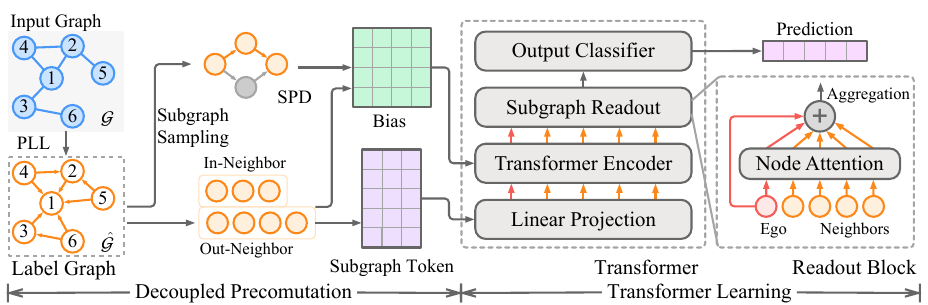}
  \caption{High-level framework of \mname{} including two consecutive stages of precomputation and Transformer learning. The precomputation stage processes input graph into the label graph, which is then used to generate the subgraph structure and SPD bias. During training, the subgraph tokens are applied as input features for each node, while SPD is regarded as positional encoding for Transformer layers. A readout block aggregates the subgraph representation to the ego node for prediction.}
\label{fig:framework}
\end{figure*}

% ====================
\section{Methodology}
\label{sec:method}
In this section, we describe the design motivation and approach of the \mname{} model by respectively elaborating on the proposed modules in its precomputation and learning stages. 
\cref{fig:framework} illustrates the overview of the \mname{} pipeline. 

{%\small 
\begin{algorithm}[bp]
\algrenewcommand{\alglinenumber}[1]{\scriptsize\bfseries#1}
% \fontsize{9pt}{12pt}\selectfont
\renewcommand{\algorithmicrequire}{\textbf{Input:}}
\renewcommand{\algorithmicensure}{\textbf{Output:}}
\caption{\mname{} Precomputation}
\label{alg:pre}
\begin{algorithmic}[1]
\Require Graph $\mathcal{G} = \langle \mathcal{V}, \mathcal{E}\rangle$, Sample size $s_{in}, s_{out}$, Sampling exponents $r_{in}, r_{out}$
\Ensure Subgraphs for all nodes $\{\mathcal{S}(v)\}$, Extended edge set $\Hat{\mathcal{E}}$ 
\State $\Hat{\mathcal{E}} \gets \varnothing$
\State Compute $\mathcal{L}$ using \cref{alg:label}
\For{$v=1$ to $n$}
    \State Add $(v, u, \delta)$ to $\Hat{\mathcal{E}}$ for all $(u, \delta) \in \mathcal{L}(v)$
    \State $\mathcal{N}_{in} \gets \{(u, \delta^{r_{in}}) \,|\, (v, \delta) \in \mathcal{L}(u)\}$ 
    \State $\mathcal{N}_{out} \gets \{(u, \delta^{r_{out}}) \,|\, (u, \delta) \in \mathcal{L}(v)\}$
    \State Sample $s_{in}$ nodes from $\mathcal{N}_{in}$ with weights $\delta^{r_{in}}$ as $\mathcal{S}_{in}$
    \State Sample $s_{out}$ nodes from $\mathcal{N}_{out}$ with weights $\delta^{r_{out}}$ as $\mathcal{S}_{out}$
    \State $\mathcal{S}(v) \gets \{v\} \cup \mathcal{S}_{in} \cup \mathcal{S}_{out}$
    \ForAll{$(u, w)$ such that $u\in\mathcal{S}(v), w\in\mathcal{S}(v)$}
        \State Compute $b(u, w)$ using \cref{eq:spd}
        \State Add $(u, w, b(u, w))$ to $\Hat{\mathcal{E}}$
    \EndFor
\EndFor\vspace{2pt}
\State \Return {$\{\mathcal{S}(v) \,|\, v\in\mathcal{V}\}$ and $\Hat{\mathcal{E}}$ }
\end{algorithmic}
\end{algorithm}
}

% ==========
\subsection{Subgraph Generation by Labeling}
\label{ssec:method_subgraph}
\noindentparagraph{Motivation: Hierarchical GT beyond adjacency.}
Canonical Graph Transformer models \cite{ying2021graphormer,wu2022nodeformer,rampavsek2022graphgps,chen2023c} generally utilize graph adjacency for composing the input sequence in graph representation learning. However, recent advances reveal that adjacency alone is insufficient to represent the implicit graph topology. GTs can be improved by modeling node connections not limited to explicit edges, and more hierarchical information benefits learning high-level knowledge on graph data \cite{zhang2022ansgt,kong2023goat,zhu2023hsgt}. 

Unlike existing hierarchical GTs relying on the original graph $\mathcal{G}$, we seek to retrieve structural information from the label graph $\Hat{\mathcal{G}}$ generated by \cref{alg:label}. As showcased in \cref{sec:pll}, the label graph hierarchy processes properties of maintaining local neighborhoods while adding global edges. This is preferable for Graph Transformers as it extends the receptive field beyond local neighbors described by graph adjacency and highlights those distant but important landmarks in the graph for attention modules on node connections. The hierarchical information is especially useful for complicated scenarios, such as heterophilous graphs, where the local graph topology may be distributive or even misleading. Moreover, the edge weight of the label graph, i.e., the shortest distance between node pairs of interest, can serve as a straightforward metric for evaluating relevance with the ego node.

To leverage the label graph efficiently, we employ a decoupling scheme to prepare the labels and necessary data in a separate stage before training. The graph data is only processed in this precomputation stage and is prevented from being fully loaded onto GPU devices, which intrinsically reduces the GPU memory overhead and offers better scalability to large graphs. 

\noindentparagraph{Sampling for Subgraph Tokens.}
\cref{alg:pre} describes the precomputation process in \mname{}. Given the input graph $\mathcal{G}$, we first build the graph labels by PLL as outlined in \cref{alg:label}. For each node, we generate a token for GT learning, which represents the neighborhood around the node in the label graph $\Hat{\mathcal{G}}$. Since the neighborhood size is variable, we convert it into a fixed-length subgraph token $\mathcal{S}(v)$ with $s$ nodes by weighted sampling, as shown in lines 4-9 in \cref{alg:pre}. Neighbors in $\mathcal{N}_{in}(v)$ and $\mathcal{N}_{out}(v)$ are sampled separately with different sizes, as \cref{ssec:pll_property} shows that these two sets contain nodes of differing importance. The distance to the ego node $b(u,v)$ is used as the sampling weight, with hyperparameters $r_{in}, r_{out} \in \RR{}$ controlling the relative importance. Note that under our sampling scheme, nodes not connected to the ego node will not appear in the token.

Overall, the subgraph generation process produces a node list of length $s = s_{in} + s_{out} + 1$ for each node, representing its neighborhood in the label graph $\Hat{\mathcal{G}}$. The relative values of hyperparameters $s_{in}$ and $s_{out}$ can be used to balance the ratio of in-neighbors and out-neighbors in $\Hat{\mathcal{G}}$, which correspond to local long-tailed nodes and distant landmark nodes in $\mathcal{G}$, respectively. Compared to canonical GT tokens representing the graph node in the context of the full graph, \mname{} only relies on a small but informative subgraph of fixed size $s$. When the graph scales up, \mname{} enjoys better scalability as its token size does not increase with the graph size.

% ==========
\subsection{Fast Subgraph Positional Encoding}
% \subsection{Hierarchical Positional Encoding and Feature}
Positional encoding is critical for GT expressivity to model inter-node relationship for graph learning. In our approach, positional encoding provides the relative identity of nodes within the subgraph hierarchy. We particularly employ shortest path distance (SPD) to token nodes as the positional encoding scheme in \mname{}, which is superior as it holds meaningful values for arbitrary node pairs regardless of locality. In comparison, other approaches such as graph proximity and eigenvectors are usually too sparse to provide identifiable information within sampled subgraphs. 

Conventionally, calculating SPD for positional encoding demands $O(n^2)$ or higher complexity as analyzed in \cref{tab:complexity}, which is not practical for large-scale scenarios. Thanks to the graph labeling computation, we are able to efficiently acquire SPD inside subgraphs. 
% Graph labeling is also useful for efficiently computing shortest distances between node pairs. 
Recalling the definition of 2-hop cover in \cref{sec:preliminaries}, we exploit the following corollary, which ensures the SPD of any node pairs can be effectively acquired on top of PLL labels: 
\vspace{-3pt}
\begin{corollary}[\cite{akiba2013}]
\label{thm:spd}
    For any node pair $(u,v)$, the shortest path distance can be calculated by:
\vspace{-3pt}
\begin{equation}
\label{eq:spd}
    b(u, v) = \min\left\{ b(u,w) + b(w,v) \,|\, w\in\mathcal{L}(u), w\in\mathcal{L}(v) \right\},
\vspace{-3pt}
\end{equation}
    where labels $\mathcal{L}$ are computed by \cref{alg:label}. Note that $b(v,v)=0$. 
\end{corollary}

% Node-pair SPDs can be utilized as positional encoding to represent the relative subgraph hierarchy in self-attention of GTs \cite{ying2021graphormer}. 
The second part of \cref{alg:pre} in line 10-12 depicts the process of further reusing the label graph data structure for managing SPD within node-wise subgraphs. For node pairs of each subgraph $\mathcal{S}(v)$, the SPDs are computed and stored as weighted edges that extend the label graph edge set $\Hat{\mathcal{E}}$. 

To employ SPD positional encoding, the transformer layer in \cref{eq:gt} is altered with a bias term $\bm{B} \in\RR{s\times s}$:
\begin{equation}
    \Tilde{\bm{H}} = \softmax\left( \frac{\bm{Q}\bm{K}^\top}{\sqrt{d_K}} + \bm{B} \right) \bm{V} ,
\label{eq:gt_bias}
\end{equation}
where the value of bias entry is a learnable parameter indexed by the node-pair SPD value $\bm{B}[u, v] = f_B(b(u,v)), f_B: \mathbb{N} \to \RR{}$. 

% ==========
\subsection{Model Architecture}
\mname{} enhances the GT architecture \cite{ying2021graphormer,zhang2022ansgt} to fit the precomputed subgraphs and mini-batch training for large-scale representation learning. Apart from the SPD bias, we also design specific modules to adapt to subgraph hierarchical learning. 
For each node $v\in\mathcal{V}$, given the subgraph $\mathcal{S}(v)$ produced by \cref{alg:pre}, input representations are retrieved from the node attributes based on the input node token as $\bm{H}^{(0)} = \text{MLP}_X(\bm{X}[\mathcal{S}(v)])$, where $\bm{X}[\mathcal{S}(v)]$ denotes node attributes $\bm{X}[u]$ for all $u\in\mathcal{S}(v)$, and $\text{MLP}_X: \RR{s\times F_0} \to \RR{s\times F}$ with hidden dimension $F$. 

For the $l$-th Transformer layer, the representation is updated as:
\begin{equation}
\begin{aligned}
    \Tilde{\bm{H}}^{(l-1)} & =\operatorname{MHA}\left(\mathrm{LN}\left(\bm{H}^{(l-1)}\right)\right)+\bm{H}^{(l-1)} , \\
    \bm{H}^{(l)} & =\operatorname{FFN}\left(\mathrm{LN}\left(\Tilde{\bm{H}}^{(l-1)}\right)\right)+\Tilde{\bm{H}}^{(l-1)} ,
\end{aligned}
\end{equation}
where $\mathrm{LN}$ and $\mathrm{FFN}$ stand for layer normalization and feed-forward network, respectively, and $\mathrm{MHA}$ denotes the multi-head self-attention architecture described by \cref{eq:attn,eq:gt_bias,eq:gt}. 

Lastly, a readout layer calculates node-wise attention over the $L$-layer representation among nodes in the fixed-length token $\mathcal{S}(v)$:
\begin{equation}
    \alpha_u = \frac{\exp\left( (\bm{H}^{(L)}[v] \| \bm{H}^{(L)}[u]) \bm{W}_E \right)}
    {\sum_{u\in\mathcal{S}(v)} \exp\left( (\bm{H}^{(L)}[v] \| \bm{H}^{(L)}[u]) \bm{W}_E \right)},
\end{equation}
which measures the correlation between ego node and its neighbors in the subgraph. The representation is then aggregated to the ego node $v$ as output:
\begin{equation}
    \bm{Z} = \text{MLP}_Z \bigg( \bm{H}^{(L)}[v] + \sum_{u\in\mathcal{S}(v)} \alpha_u \bm{H}^{(L)}[u] \bigg),
\end{equation}
where $\text{MLP}_Z$ is the output classifier. 

\noindentparagraph{Virtual Node.}
We add virtual nodes representing landmarks to $\Hat{\mathcal{G}}$ such that $(0, v) \in \Hat{\mathcal{E}}, b(0,v) = \infty$ for all nodes $v\in\mathcal{V}$. It can be observed from \cref{alg:label} that virtual nodes are added to every $\mathcal{S}(v)$ without affecting label construction and SPD query. During the learning stage, we set their attributes and attention bias to be learnable. This scheme actually generalizes the global virtual node utilized in \cite{ying2021graphormer}, offering graph-level context to node-level representation during representation updates. 

\noindentparagraph{Mini-batch Capability.}
Remarkably, throughout the Transformer learning stage of \mname{}, input data including subgraph tokens, SPD bias, and node attributes are all readily prepared by \cref{alg:pre} as described in previous subsections. For each node, only indexing operations are performed on $\bm{X}$ and $\Hat{\mathcal{E}}$ based on the subgraph token $\mathcal{S}(v)$, and no graph-scale computation is required during learning iterations. Therefore, mini-batch training for \mname{} can be easily implemented by sampling batches of ego nodes, and only indexed strides of $\bm{X}$ and $\Hat{\mathcal{E}}$ are loaded onto GPU devices.

\begin{table*}[!t]
% \vspace{1ex}
% \captionsetup{skip=6pt,font={small}}
\begin{adjustbox}{max width=\textwidth}
\begin{threeparttable}
\caption{Effectiveness and efficiency results on heterophilous datasets, while evaluation on homophilous datasets are in \cref{tab:resapp}. 
``Pre.'' , ``Epoch'', and ``Infer'' are precomputation, training epoch, and inference time (in seconds), respectively. ``Mem.'' refers to peak GPU memory throughout the whole learning process ($\si{GB}$). Respective results of the first and second best performances in each dataset are marked in \textbf{bold} and {underlined} fonts. 
}
\label{tab:res}
\centering
\vspace{-1ex}
% \small
% \fontsize{8pt}{6pt}\selectfont
\setlength{\tabcolsep}{3pt}
\renewcommand{\arraystretch}{1.05}
\newcommand{\szs}[1]{{\fontsize{8pt}{6pt}\selectfont #1}}
\newcommand{\dsheader}[2]{\multicolumn{5}{#1}{\ds{#2}}}
\begin{tabular}{@{}c|ccccc|ccccc|ccccc@{}}
\toprule
  \multirow{2}{*}{\textbf{Small}} & 
    \dsheader{c|}{chameleon} & \dsheader{c|}{squirrel} & \dsheader{c}{tolokers} \\ 
  ~ & \szs{Pre.} & \szs{Epoch} & \szs{Infer} & \szs{Mem.} & \szs{Acc} 
    & \szs{Pre.} & \szs{Epoch} & \szs{Infer} & \szs{Mem.} & \szs{Acc} 
    & \szs{Pre.} & \szs{Epoch} & \szs{Infer} & \szs{Mem.} & \szs{ROC AUC}   \\
\midrule
  DIFFormer$^\ast$ & - & 0.09 & 0.38 & 0.50 & 37.83\tpm{4.54} & - & 0.05 & 0.05 & 0.7 & 35.73\tpm{1.37} & - & 0.16 & 85.8 & 0.88 & 74.88\tpm{0.59} \\
  PolyNormer$^\ast$ & - & 0.03 & 0.17 & 1.1 & 40.70\tpm{3.38} & - & 0.07 & 0.49 & 1.2 & \textbf{38.40}\tpm{1.10} & - & 1.27 & 15.5 & 9.4 & 79.39\tpm{0.50} \\
  NAGphormer & 0.27 & 0.03 & 0.03 & 0.5 & 33.18\tpm{4.30} & 0.85 & 0.08 & 0.08 & 0.5 & 32.02\tpm{3.93} & 1.59 & 0.11 & 0.02 & 0.5 & 79.32\tpm{0.39}\\
  ANS-GT & 11.2 & 1.98 & 0.78 & 2.8 & \ul{41.19}\tpm{0.69} & 28.1 & 4.48 & 1.95 & 6.6 & 37.15\tpm{1.10} & 716 & 2.37 & 3.42 & 10.7 & 79.31\tpm{0.97}\\
  GOAT & 1.99 & 0.34 & 0.44 & 0.4 & 35.02\tpm{1.15} & 6.66 & 0.37 & 0.58 & 0.6 & 30.78\tpm{0.91} & 36.1 & 5.49 & 5.87 & 5.0 & \ul{79.46}\tpm{0.57}\\
  HSGT$^\ast$ & 0.01 & 0.34 & 0.73 & 0.3 & 32.28\tpm{2.43} & 0.01 & 0.42 & 0.74 & 0.4 & 34.32\tpm{0.51} & 2.62 & 7.76 & 8.12 & 17.4 & 79.24\tpm{0.83}\\
  \textbf{\mname{} (ours)} & 0.08 & 0.03 & 0.005 & 4.5 & \textbf{43.63}\tpm{2.34} & 0.35 & 0.68 & 0.01 & 5.7 & \ul{37.16}\tpm{0.57} & 1.9 & 0.17 & 0.02 & 7.2 & \textbf{79.86}\tpm{0.47} \\[3pt]

\toprule
  \multirow{2}{*}{\textbf{Large}} & 
    \dsheader{c|}{penn94} & \dsheader{c|}{genius} & \dsheader{c}{twitch-gamer} \\ 
  ~ & \szs{Pre.} & \szs{Epoch} & \szs{Infer} & \szs{Mem.} & \szs{Acc} 
    & \szs{Pre.} & \szs{Epoch} & \szs{Infer} & \szs{Mem.} & \szs{Acc} 
    & \szs{Pre.} & \szs{Epoch} & \szs{Infer} & \szs{Mem.} & \szs{Acc}   \\
\midrule
  DIFFormer$^\ast$ & - & 0.53 & 0.65 & 5.5 & 61.77\tpm{3.41} & - & 0.77 & 5.47 & 5.4 & 84.52\tpm{0.36} & - & 0.61 & 5.14 & 4.9 & 60.81\tpm{0.44} \\
  PolyNormer$^\ast$ & - & 0.58 & 18.4 & 6.3 & \textbf{79.87}\tpm{0.06} & - & 0.77 & 28 & 12.9 & \ul{85.64}\tpm{0.52} & - & 1.45 & 89 & 21.4 & \ul{64.72}\tpm{0.65} \\
  NAGphormer & 237 & 6.14 & 2.13 & 2.3 & 74.45\tpm{0.60} & 38 & 5.43 & 1.04 & 2.3 & 83.88\tpm{0.13} & 16 & 1.92 & 0.39 & 2.3 & 61.92\tpm{0.19}\\
  ANS-GT & 3889 & 42 & 4.9 & 8.7 & 67.76\tpm{1.32} & 34092 & 37 & 4.95 & 8.7 & 67.76\tpm{1.32} & 12924 & 19 & 6.7 & 8.6 & 61.55\tpm{0.45}\\
  GOAT & 1332 & 33 & 18 & 20.9 & 71.42\tpm{0.44} & 2664 & 28 & 39 & 8.9 & 80.12\tpm{2.32} & 3348 & 37 & 63 & 21.2 & 61.38\tpm{0.83}\\
  HSGT$^\ast$ & 12 & 115 & 110 & 9.3 & 67.77\tpm{0.27} & 21 & 98 & 114 & 17.1 & 84.03\tpm{0.24} & 68 & 235 & 253 & 11.2 & 61.60\tpm{0.09}\\
  \textbf{\mname{} (ours)} & 31 & 14 & 0.3 & 10.2 & \ul{78.74}\tpm{0.45} & 52 & 5.4 & 0.33 & 7.0 & \textbf{91.06}\tpm{0.47} & 172 & 2.2 & 0.15 & 7.3 & \textbf{67.03}\tpm{2.17} \\
\bottomrule
\end{tabular}
\begin{tablenotes}
\small
    \item [$\ast$] Inference of these models is performed on the CPU in a full-batch manner due to their requirement of the whole graph.
\end{tablenotes}
\end{threeparttable}
\end{adjustbox}
\vspace{-.8ex}
\end{table*}

% ==========
\subsection{Complexity Analysis}
To characterize the model scalability, we consider the time and memory complexity of \mname{} separately in the precomputation and learning stages. In precomputation, the PLL labeling and sampling process \cref{alg:label} satisfies the analysis in \cite{akiba2013}, entailing a complexity of $O(ns+ms)$ for computing labels of all nodes. 
Regarding the positional encoding, a single SPD query following \cref{thm:spd} can be calculated in $O(s)$ time within the subgraph. The query is performed at most $O(ns^2)$ times for all nodes, which leads to an $O(ns^3)$ overhead for $\Hat{\mathcal{E}}$ in total. It is worth noting that the empirical number of queries is significantly smaller than the above bound, since the subgraphs $\mathcal{S}(v)$ are highly overlapped for neighboring nodes. 
The memory overhead for managing sampled tokens and features in RAM is $O(ns^2)$ and $O(nF)$, respectively. Note that SPD values are stored as integers, which is more efficient than other positional encoding schemes.  

During model training, one epoch of $L$-layer feature transformation on all nodes demands $O(LnF)$ operations, while bias projection is performed with $O(ns^2)$ time complexity. The GPU memory footprint for handling a batch of node representations and bias matrices is $O(Ln_bF)$ and $n_bs^2$, respectively, where $n_b$ is the batch size. It can be observed that the training overhead is only determined by batch size and is free from the graph scale, ensuring favorable scalability for iterative GT updates.

% \blue{Precompute: (1) indexing: . (2) feature generation: $O(ns^2*|\mathcal{L}|_{avg}$)=$O(ns^3)$?}

% !TEX root = ../main.tex

\section{Experiments}
\label{sec:experiments}
We comprehensively evaluate the performance of \mname{} with a wide range of datasets and baselines. In \cref{ssec:exp_main}, we highlight the model efficiency regarding time and memory overhead, as well as its effectiveness under both homophily and heterophily. \cref{ssec:exp_param,ssec:exp_ablation} provides in-depth insights into the effect of \mname{} designs in exploiting graph hierarchy. Implementation details and full experimental results can be found in \cref{seca:exp}.

% ==========
\vspace{-2pt}
\subsection{Experimental Settings}
\noindentparagraph{Tasks and Datasets.}
We focus on the node classification task on 12 benchmark datasets in total covering both homophily \cite{Sen2008,shchur2018pitfalls,Hu2020} and heterophily \cite{lim2021,olegplatonov2023}, whose statistics are listed in \cref{tab:dataset}. Compared to conventional graph learning tasks used in GT studies, this task requires learning on large single graphs, which is suitable for assessing model scalability. We follow common data processing and evaluation protocols as detailed in \cref{sseca:exp_settings}. 
Evaluation is conducted on a server with 32 Intel Xeon CPUs (2.4GHz), an Nvidia A30 GPU (24GB memory), and 512GB RAM. 

\noindentparagraph{Baselines.}
Since the scope of this work lies in the efficacy and efficiency enhancement of the GT architecture, we primarily compare against state-of-the-art Graph Transformer models with attention-based layers and mini-batch capability. Methods including DIFFormer \cite{wu2023difformer} and PolyNormer \cite{deng2024polynormer} are considered as kernel-based approaches. NAGphormer \cite{chen2023c}, GOAT \cite{kong2023goat}, HSGT \cite{zhu2023hsgt}, and ANS-GT \cite{zhang2022ansgt} stand for hierarchical GTs. 

\noindentparagraph{Evaluation Metrics.}
We use ROC AUC as the efficacy metric on \ds{tolokers} and classification accuracy on the other datasets. For efficiency evaluation, we notice that there is limited consensus due to the great variety in GT training schemes. Therefore, we attempt to employ a comprehensive evaluation considering both time and memory overhead for a fair comparison. Model speed is represented by the average training time per epoch and the inference time on the testing set. For models with graph precomputation, the time for this process is separately recorded. We also feature the GPU memory footprint, which is the scalability bottleneck.

% ==========
\vspace{-5pt}
\subsection{Performance Comparison}
\label{ssec:exp_main}
\cref{tab:res} presents the efficacy and efficiency evaluation results on 6 heterophilous datasets, while metrics for 6 homophilous graphs can be found in \cref{tab:resapp}. 
As an overview, \mname{} demonstrates fast computation speed and favorable mini-batch scalability throughout the learning process. It also reaches top-tier accuracy by outperforming the state-of-the-art GTs on multiple datasets.

\noindentparagraph{Time Efficiency.}
Benefiting from the decoupled architecture, \mname{} is powerful in achieving competitive speed with existing efficiency-oriented GT designs. For baselines with heavy precomputation overhead, including ANS-GT and GOAT, \mname{} showcases speed improvements by orders of magnitude, with up to $700\times$ boost over ANS-GT on \ds{genius}. Aligned with our complexity analysis in \cref{sec:related}, the key impact factor of \mname{} is the node size $n$ and is less affected by $m$ and $F$ compared to precomputation in other methods. 
Meanwhile, \mname{} is capable of performing the fastest inference even on large-scale graphs, thanks to its simple model transformation without graph-scale operations. Its training speed is also on par with the best competitors, which usually exploit highly simplified architectures. In contrast, models including PolyNormer and HSGT suffer from longer learning times due to their iterative graph extraction and transformation.

\noindentparagraph{Memory Footprint.}
In modern computing platforms, GPU memory is usually highly constrained and becomes the scalability bottleneck for the resource-intensive graph learning. \mname{} exhibits efficient utilization of GPU for training with larger batch sizes while avoiding the out-of-memory issue. In comparison, drawbacks in several model designs prevent them from efficiently performing GPU computation, which stems from the adoption of graph operations. Notably, kernel-based models require full graph message-passing in their inference stage, which is largely prohibitive on GPUs and can only be conducted on CPUs. HSGT faces the similar issue caused by its graph coarsening module. We note that these solutions are less scalable and hinder the GPU utilization during training. 
In addition, ANS-GT typically demands high memory footprint for storing and adjusting its subgraphs, which exceeds the memory limit of our platform in \cref{tab:resapp}.

\noindentparagraph{Prediction Accuracy.}
\mname{} successfully achieves significant accuracy improvement on several heterophily datasets such as \ds{chameleon} and \ds{twitch-gamer}, while the performance on other heterophilous and homophilous datasets in \cref{tab:res,tab:resapp} is also comparable with the state of the art. We attribute the performance gain to the application of the label graph hierarchy in \mname{}, which effectively addresses the heterophily issue of these graphs as analyzed in \cref{sec:pll}. Since the label graph also preserves edges in the raw graph, the performance of \mname{} is usually not lower than learning on the latter. 
In comparison, baseline methods without heterophily-oriented designs, including DIFFormer, NAGphormer, and HSGT, perform generally worse on these graphs. This is because their models tend to rely on the raw adjacency or even promote it with higher modularity. As a consequence, node connections retrieved by GT attention modules are restrained in the local neighborhood and fail to produce accurate classifications. On the other hand, while PolyNormer achieves remarkable accuracy on several heterophilous graphs thanks to its strong expressivity, its performance is largely suboptimal on homophilous graphs in \cref{tab:resapp}.

\begin{figure}[tp]
\captionsetup[subfigure]{skip=4pt}
\centering
    \subcaptionbox{Sample Size \label{fig:param_s_cha}}%
    {\includegraphics[width=0.44\columnwidth]{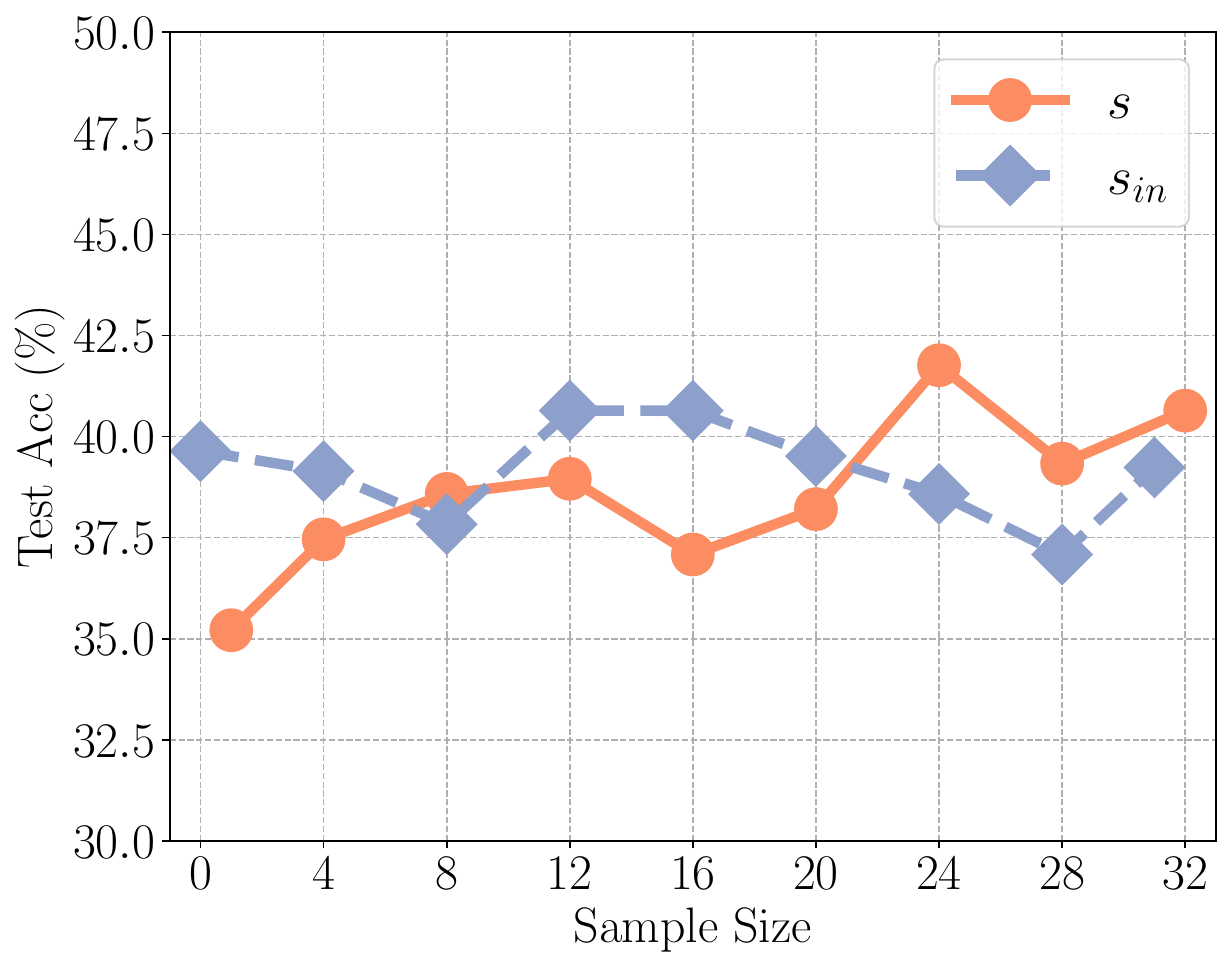}}
\hfil
    \subcaptionbox{Sampling Weight \label{fig:param_r_cha}}%
    {\includegraphics[width=0.44\columnwidth]{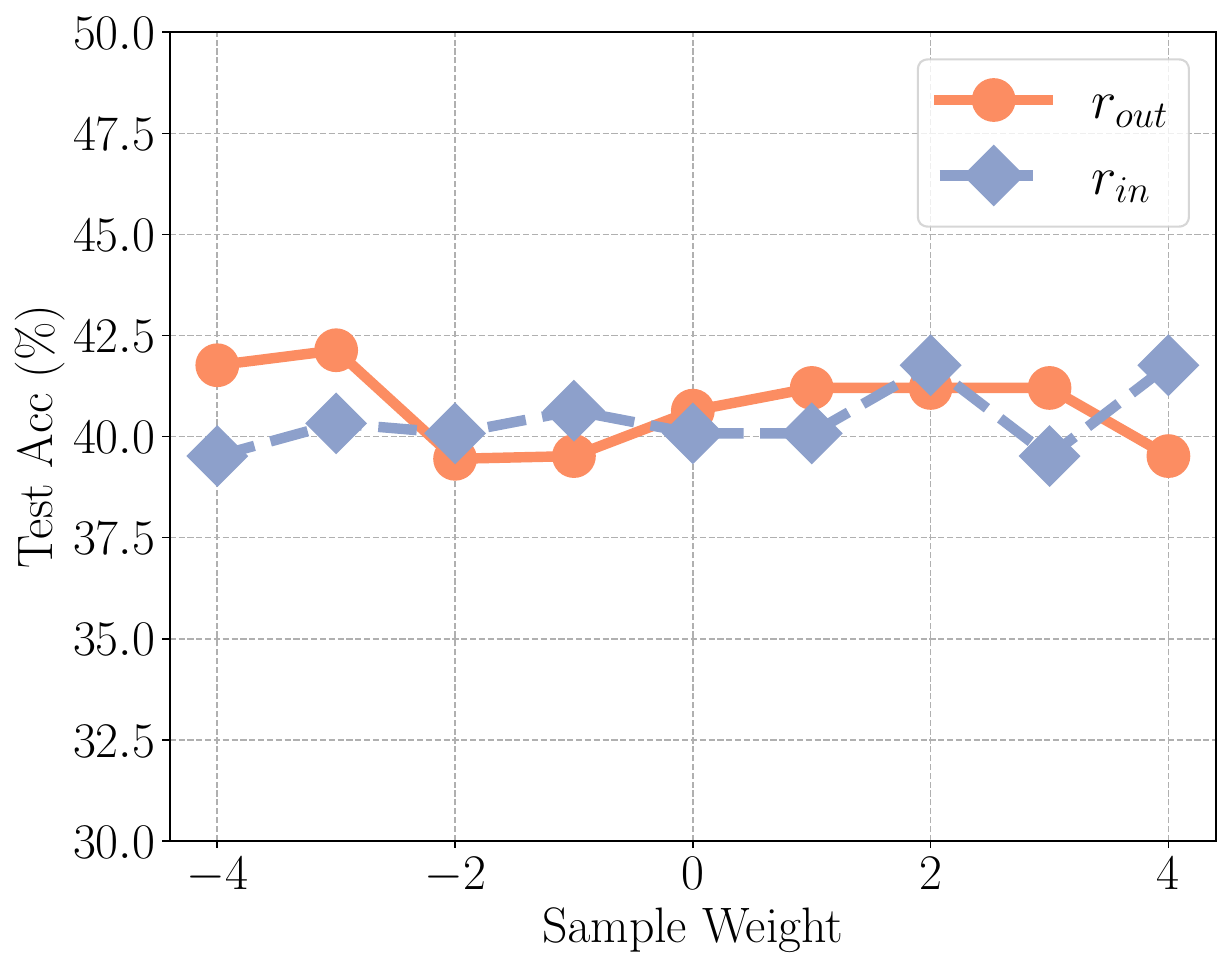}}
\caption{Effect of subgraph parameters on \ds{chameleon}. }
\label{fig:param}
\vspace{-1.0em}
\end{figure}

% ==========
\subsection{Effect of Hyperparameters}
\label{ssec:exp_param}
We then study the effectiveness of the label graph hierarchy in \mname{} featuring the subgraph generation process in \cref{fig:param}, which displays the
impact of sample size $s, s_{in}$ and sampling exponents $r_{in}, r_{out}$ corresponding to \cref{alg:pre}. 
Regarding the total subgraph size $s$, it can be observed from \cref{fig:param_s_cha} that a reasonably large $s$ is essential for effectively representing graph labels and achieving stable accuracy. In the main experiments, we uniformly adopt a constant $s=32$ token size across all datasets, as it is large enough to cover the neighborhood of most nodes while maintaining computational efficiency. As a reference, the average neighborhood size among all nodes is 16.0 on \ds{citeseer} and 31.2 on \ds{chameleon}. 
Within the fixed token length, an equal partition for in- and out-neighbors is preferable according to \cref{fig:param_s_cha}, where impact of $s_{in}$ is shown when $s=32$. 

\cref{fig:param_r_cha} presents the result of changing the sampling weight factor. For \ds{chameleon}, in the plot, negative exponents favoring nodes with small SPD values are more advantageous to the model performance. Nonetheless, the variance is not significant as long as the subgraph effectively covers the neighborhood of the majority of nodes. We hence conclude that \cref{alg:pre} for \mname{} precomputation does not require precise hyperparameter tuning.

% ==========
\subsection{Ablation Study}
\label{ssec:exp_ablation}
\cref{tab:ablation} examines the respective effectiveness of the hierarchical modules in the \mname{} network architecture, where we separately present results on homophilous and heterophilous datasets. It can be observed that the model without SPD bias suffers the greatest accuracy drop, since topological information represented by positional encoding is necessary for GTs to retrieve the relative connection between nodes and gain performance improvement over learning plain node-level features.

In \mname{}, the learnable virtual node representation is invoked to provide adaptive graph-level context before Transformer layers, while the attention-based node-wise readout module aims to distinguish nodes inside subgraphs and aggregate useful representation after encoder transformation. As shown in \cref{tab:ablation}, both modules achieve relatively higher accuracy improvements on the heterophilous graph \ds{chameleon}, which validates that the proposed designs are particularly suitable for addressing the heterophily issue by recognizing hierarchical information.

\begin{table}[h]
\vspace{-1ex}
\caption{Ablation study of \mname{} model components. The first line shows the accuracy of the complete \mname{} architecture. Each subsequent line indicates the performance difference when the specified module is removed.}
\label{tab:ablation}
\centering
% \small
% \setlength{\tabcolsep}{2.5pt}
\begin{adjustbox}{max width=\columnwidth}
\begin{tabular}{c|cc|cc} \toprule
    \textbf{Dataset} & \ds{citeseer} & $\Delta$ & \ds{chameleon} & $\Delta$ \\ 
\midrule
    \mname{} & 74.91 & -- & 43.63 & -- \\
    $-$ Node Readout & 72.21 & -2.70 & 38.76 & -4.87 \\
    $-$ Virtual Node & 71.15 & -3.76 & 37.08 & -6.55 \\
    $-$ SPD Bias & 68.55 & -6.36 & 36.52 & -7.11 \\
\bottomrule
\end{tabular}
\end{adjustbox}
\vspace{-1em}
\end{table}

% !TEX root = ../main.tex

\section{Conclusion}
\label{sec:conclusion}
In this work, we present \mname{} for leveraging decoupled graph hierarchy by graph labeling. Our analysis reveals that the label graph exhibits an informative hierarchy and enhances attention learning on the connections between nodes. Regarding efficiency, construction and distance query of the label graph can be accomplished with \textit{linear} complexity and are decoupled from iterative model training. Hence, the model benefits from scalability in computation speed and mini-batch training. Empirical evaluation showcases the superiority of \mname{} especially under heterophily.

\clearpage
% \balance
\begin{acks}

\end{acks}

\begin{table*}[!b]
% \vspace{1ex}
% \captionsetup{skip=6pt,font={small}}
\begin{adjustbox}{max width=\textwidth}
\begin{threeparttable}
\caption{Effectiveness and efficiency results on homophilous datasets. 
``Pre.'' , ``Epoch'', and ``Infer'' are precomputation, training epoch, and inference time (in seconds), respectively. ``Mem.'' refers to peak GPU memory throughout the whole learning process ($\si{GB}$). ``$\si{OOM}$'' stands for out of memory error. 
Respective results of the first and second best performances on each dataset are marked in \textbf{bold} and {underlined} fonts. 
}
\label{tab:resapp}
\centering
% \small
% \fontsize{8pt}{6pt}\selectfont
\setlength{\tabcolsep}{3pt}
\renewcommand{\arraystretch}{1.05}
\newcommand{\szs}[1]{{\fontsize{8pt}{6pt}\selectfont #1}}
\newcommand{\dsheader}[2]{\multicolumn{5}{#1}{\ds{#2}}}
\begin{tabular}{@{}c|ccccc|ccccc|ccccc@{}}
\toprule
  \multirow{2}{*}{\textbf{Small}} & 
    % \dsheader{c}{reddit|} & \dsheader{c|}{ogbn-arxiv} & \dsheader{c}{ogbn-mag} \\ 
    \dsheader{c|}{cora} & \dsheader{c|}{citeseer} & \dsheader{c}{pubmed} \\ 
  ~ & \szs{Pre.} & \szs{Epoch} & \szs{Infer} & \szs{Mem.} & \szs{Acc} 
    & \szs{Pre.} & \szs{Epoch} & \szs{Infer} & \szs{Mem.} & \szs{Acc} 
    & \szs{Pre.} & \szs{Epoch} & \szs{Infer} & \szs{Mem.} & \szs{Acc}   \\
\midrule
  DIFFormer$^\ast$ & - & 0.11 & 0.13 & 1.2 & 83.37\tpm{0.50} & - & 0.07 & 0.07 & 1.7 & \ul{74.65}\tpm{0.67} & - & 0.37 & 0.35 & 2.7 & 75.77\tpm{0.40} \\
  PolyNormer$^\ast$ & - & 0.11 & 0.65 & 1.4 & 80.43\tpm{1.55} & - & 0.21 & 0.86 & 1.6 & 68.70\tpm{0.95} & - & 0.86 & 6.07 & 2.5 & 75.80\tpm{0.46} \\
  NAGphormer & 0.68 & 0.01 & 0.06 & 0.5 & 76.96\tpm{0.73} & 1.26 & 0.01 & 0.38 & 0.5 & 62.26\tpm{2.10} & 3.05 & 0.01 & 0.04 & 0.5 & 78.46\tpm{1.01}\\
  ANS-GT & 43 & 2.0 & 1.12 & 2.0 & \ul{85.42}\tpm{0.52} & 59.9 & 11.65 & 4.25 & 11.9 & 73.58\tpm{0.98} & 529 & 14 & 3.52 & 1.9 & \ul{89.53}\tpm{0.51}\\
  GOAT & 10.1 & 0.25 & 0.93 & 2.5 & 78.26\tpm{0.17} & 11.1 & 0.31 & 1.04 & 2.1 & 64.69\tpm{0.43} & 57.4 & 0.34 & 1.61 & 5.3 & 77.76\tpm{0.97} \\
  HSGT$^\ast$ & 0.1 & 1.81 & 2.33 & 0.5 & 81.73\tpm{1.95} & 0.06 & 0.87 & 1.23 & 0.9 & 69.72\tpm{1.02} & 5.0 & 3.89 & 4.44 & 24 & 88.86\tpm{0.46} \\
  \textbf{\mname{} (ours)} & 0.42 & 0.05 & 0.005 & 10.1 & \textbf{85.58}\tpm{0.18} & 0.43 & 0.05 & 0.006 & 9.6 & \textbf{74.91}\tpm{0.64} & 2.6 & 0.25 & 0.05 & 9.4 & \textbf{89.80}\tpm{0.48} \\
  
\toprule
\multirow{2}{*}{\textbf{Large}} & 
    \dsheader{c|}{physics} & \dsheader{c|}{ogbn-arxiv} & \dsheader{c}{ogbn-mag} \\ 
    % \dsheader{c|}{ogbn-arxiv} & \dsheader{c|}{ogbn-mag} & \dsheader{c}{ogbn-products} \\ 
  ~ & \szs{Pre.} & \szs{Epoch} & \szs{Infer} & \szs{Mem.} & \szs{Acc} 
    & \szs{Pre.} & \szs{Epoch} & \szs{Infer} & \szs{Mem.} & \szs{Acc} 
    & \szs{Pre.} & \szs{Epoch} & \szs{Infer} & \szs{Mem.} & \szs{Acc}   \\
\midrule
  DIFFormer$^\ast$ & - & 1.73 & 3.79 & 3.3 & 96.10\tpm{0.11} & - & 0.89 & 4.1 & 2.3 & 55.90\tpm{8.23} & - & 1.72 & 9.71 & 4.2 & 31.13\tpm{0.48} \\
  PolyNormer$^\ast$ & - & 0.76 & 2.44 & 4.1 & \textbf{96.59}\tpm{0.16} & - & 0.83 & 11 & 7.2 & 73.24\tpm{0.13} & - & 20 & 992 & 22.3 & 32.42\tpm{0.15} \\
  NAGphormer & 33 & 8.43 & 2.43 & 1.1 & \ul{96.52}\tpm{0.24} & 18 & 4.36 & 0.79 & 2.3 & 67.85\tpm{0.17} & 89 & 10.3 & 2.21 & 3.8 & 33.23\tpm{0.06} \\
  ANS-GT & 2203 & 63.1 & 34.6 & 12.5 & 96.31\tpm{0.28} & 16205 & 109 & 2.72 & 11.3 & 71.06\tpm{0.48} & - & - & - & OOM & - \\
  GOAT & 45 & 13.7 & 12.2 & 8.7 & 96.24\tpm{0.15} & 1823 & 48 & 61 & 6.5 & \textbf{69.66}\tpm{0.73} & 2673 & 116 & 102 & 6.1 & 27.69\tpm{1.32}\\
  HSGT$^\ast$ & 12 & 40.7 & 61.5 & 5.0 & 96.05\tpm{0.50} & 16 & 475 & 142 & 0.3 & 68.30\tpm{0.32} & 182 & 582 & 629 & 12.6 & \ul{33.51}\tpm{1.15} \\
  \textbf{\mname{} (ours)} & 17 & 0.48 & 0.03 & 13.3 & 96.31\tpm{0.42} & 64 & 2.02 & 0.18 & 5.5 & \ul{69.17}\tpm{0.33} & 7739 & 14.5 & 0.16 & 6.7 & \textbf{33.74}\tpm{0.24} \\
  
\bottomrule
\end{tabular}
\begin{tablenotes}
\small
    \item [$\ast$] Inference of these models is performed on the CPU in a full-batch manner due to their requirement of the whole graph.
\end{tablenotes}
\end{threeparttable}
\end{adjustbox}
% \vspace{-.8ex}
\end{table*}

\iffalse
  DIFFormer$^\ast$ & - & 0.89 & 4.1 & 2.3 & 55.90\tpm{8.23} & - & 2.36 & 21 & 8.2 & 94.96\tpm{0.37} & - & 1.72 & 9.71 & 4.2 & 31.13\tpm{0.48} \\
  PolyNormer$^\ast$ & - & 0.83 & 11 & 7.2 & 73.24\tpm{0.13} & - & 5.28 & 246 & 13.5 & 96.64\tpm{0.07} & - & 20 & 992 & 22.3 & 32.42\tpm{0.15} \\
  NAGphormer & 18 & 4.36 & 0.79 & 2.3 & 67.85\tpm{0.17} & 280 & 3.23 & 0.86 & 8.9 & 95.77\tpm{0.08} & 89 & 10.3 & 2.21 & 3.8 & 33.23\tpm{0.06} \\
  ANS-GT & 16205 & 109 & 2.72 & 11.3 & 71.06\tpm{0.48} & - & - & - & OOM & - & - & - & - & OOM & - \\
  GOAT & 1823 & 48 & 61 & 6.5 & 69.66\tpm{0.73} & 628 & 141 & 104 & 16.1 & 95.52\tpm{0.52} & 2673 & 116 & 102 & 6.1 & 27.69\tpm{1.32}\\
  HSGT$^\ast$ & 16 & 475 & 142 & 0.35 & 68.30\tpm{0.32} & 614 & 453 & 482 & 8.7 & 95.12\tpm{0.17} & 182 & 582 & 629 & 12.6 & \ul{33.51}\tpm{1.15} \\
\fi

\bibliographystyle{ACM-Reference-Format}
\bibliography{main}

% \clearpage
\appendix
% !TEX root = ../main.tex

% ====================
\section{Additional Experiments}
\label{seca:exp}

% ==========
\subsection{Detailed Experiment Settings}
\label{sseca:exp_settings}
\noindentparagraph{Dataset Details.}
\cref{tab:dataset} displays the scales and heterophily status of graph datasets utilized in our work. Undirected edges twice in the table. \ds{chameleon} and \ds{squirrel} are the filtered version from \cite{olegplatonov2023}, while \ds{ogbn-mag} is the homogeneous variant. 
We employ 60/20/20 random data splitting percentages for training, validation, and testing sets, respectively, except for \ds{ogbn-mag}, where the original split is used. Regarding efficacy metrics, ROC AUC is used on \ds{tolokers} following the original settings, and accuracy is used for the rest. 

\noindentparagraph{Hyperparameters.}
Parameters regarding the precomputation stage for graph structures are discussed in \cref{ssec:exp_param}. For subgraph sampling, we perform parameter search for relative ratio of in/out neighbors represented by $s_{out}$ in rage $[0, 32]$. For sampling weights $r_{in}, r_{out}$, we search their values in range $[-4, 4]$. 

For network architectural hyperparameters, we use $L=4$ Transformer layers with $N_H = 8$ heads and $F = 128$ hidden dimension for our \mname{} model across all experiments. The dropout rates for inputs (features and bias) and intermediate representation are 0.1 and 0.5, respectively. The AdamW optimizer is used with a learning rate of $10^{-4}$. The model is trained with 500 epochs with early stopping. 
Since baseline GTs employ different batching strategies, it is difficult to unify the batch size across all models. We set the batch size to the largest value in the available range without incurring out of memory exception on our $24\si{GB}$ GPU, intending for a fair efficiency evaluation considering both learning speed and space.  

\begin{table}[ht]
\caption{Statistics of graph datasets. $f$ and $N_c$ are the numbers of input attributes and label classes, respectively. ``Train'' is the portion of training set w.r.t. labeled nodes. }
\label{tab:dataset}
\centering
% \small
\setlength{\tabcolsep}{2.5pt}
\begin{adjustbox}{max width=\columnwidth}
\begin{tabular}{c@{}c|rrrr r} \toprule
    \textbf{Hetero.} & \textbf{Dataset} & \textbf{Nodes} $n$ & \textbf{Edges} $m$
        & $F$ & $N_c$ & \textbf{Train} \\ 
\midrule \multirow{6}{*}{Homo.} 
    & \ds{chameleon} & $890$ & $17,708$ 
        & $2325$ & $5$ & $60\%$ \\
    & \ds{squirrel} & $2,223$ & $93,996$ 
        & $2089$ & $5$ & $60\%$ \\
    & \ds{tolokers} & $11,758$ & $1,038,000$ 
        & $10$ & $2$ & $60\%$ \\
    & \ds{penn94} & $41,554$ & $2,724,458$ 
        & $4814$ & $2$ & $60\%$ \\
    & \ds{genius} & $421,961$ & $1,845,736$ 
        & $12$ & $2$ & $60\%$ \\
    & \ds{twitch-gamer} & $168,114$ & $13,595,114$ 
        & $7$ & $2$ & $60\%$ \\
\midrule \multirow{6}{*}{Hetero.} 
    & \ds{cora} & $2,708$ & $10,556$ 
        & $1433$ & $7$ & $60\%$ \\
    & \ds{citeseer} & $3,279$ & $9,104$ 
        & $3703$ & $6$ & $60\%$  \\
    & \ds{pubmed} & $19,717$ & $88,648$ 
        & $500$ & $3$ & $60\%$ \\
    % \ds{cs} & & $60\%$ & homo. \\
    & \ds{physics} & $34,493$ & $495,924$ & 8415 & 5 & $60\%$ \\
    & \ds{ogbn-arxiv} & $169,343$ & $2,315,598$ 
        & $128$ & $40$ & $54\%$ \\
    % & \ds{reddit} & $232,965$ & $114,615,892$ 
    %     & $602$ & $41$ & $60\%$ \\
    & \ds{ogbn-mag} & $736,389$ & $10,792,672$ 
        & $128$ & $349$ & $85\%$ \\
\bottomrule
\end{tabular}
\end{adjustbox}
\end{table}

% \clearpage
\begin{figure*}[p]
\centering
    \subcaptionbox{\ds{cora} Raw Graph\label{fig:comm_cora}}%
    {\includegraphics[width=0.36\textwidth]{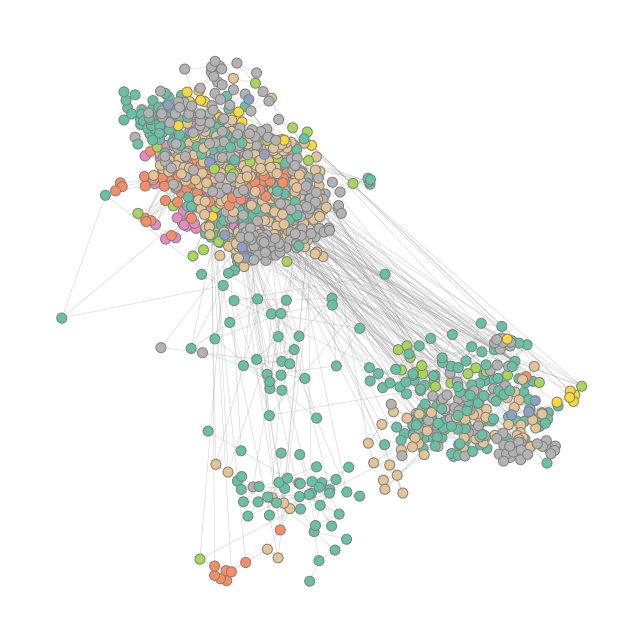}}
\hfil
    \subcaptionbox{\ds{cora} Label Graph\label{fig:colb_cora}}%
    {\includegraphics[width=0.36\textwidth]{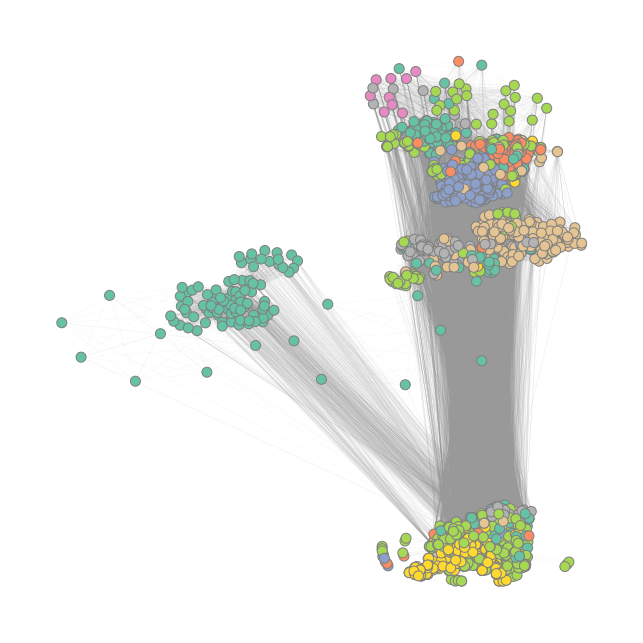}}
\\
    \subcaptionbox{\ds{citeseer} Raw Graph\label{fig:comm_citeseer}}%
    {\includegraphics[width=0.36\textwidth]{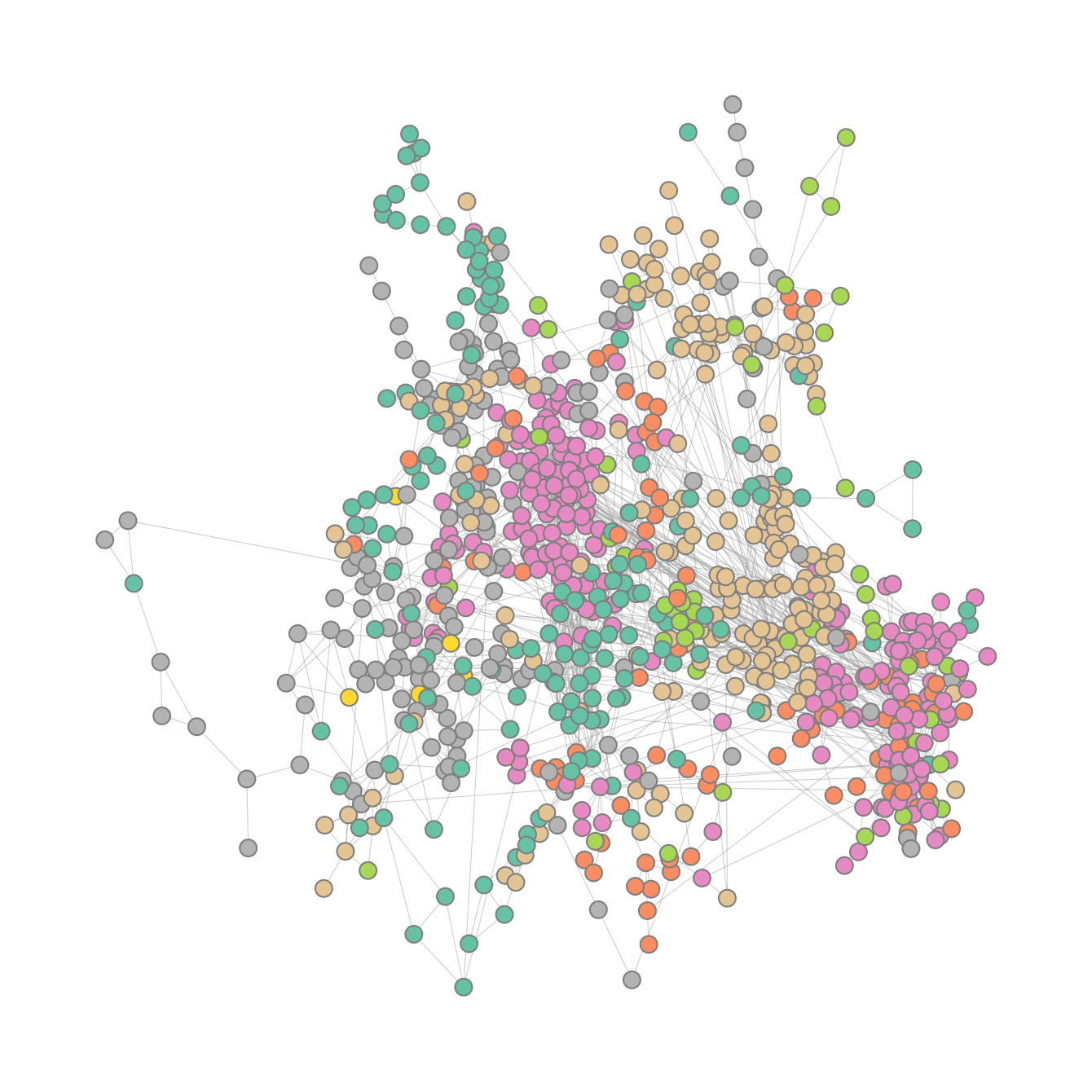}}
\hfil
    \subcaptionbox{\ds{citeseer} Label Graph\label{fig:colb_citeseer}}%
    {\includegraphics[width=0.36\textwidth]{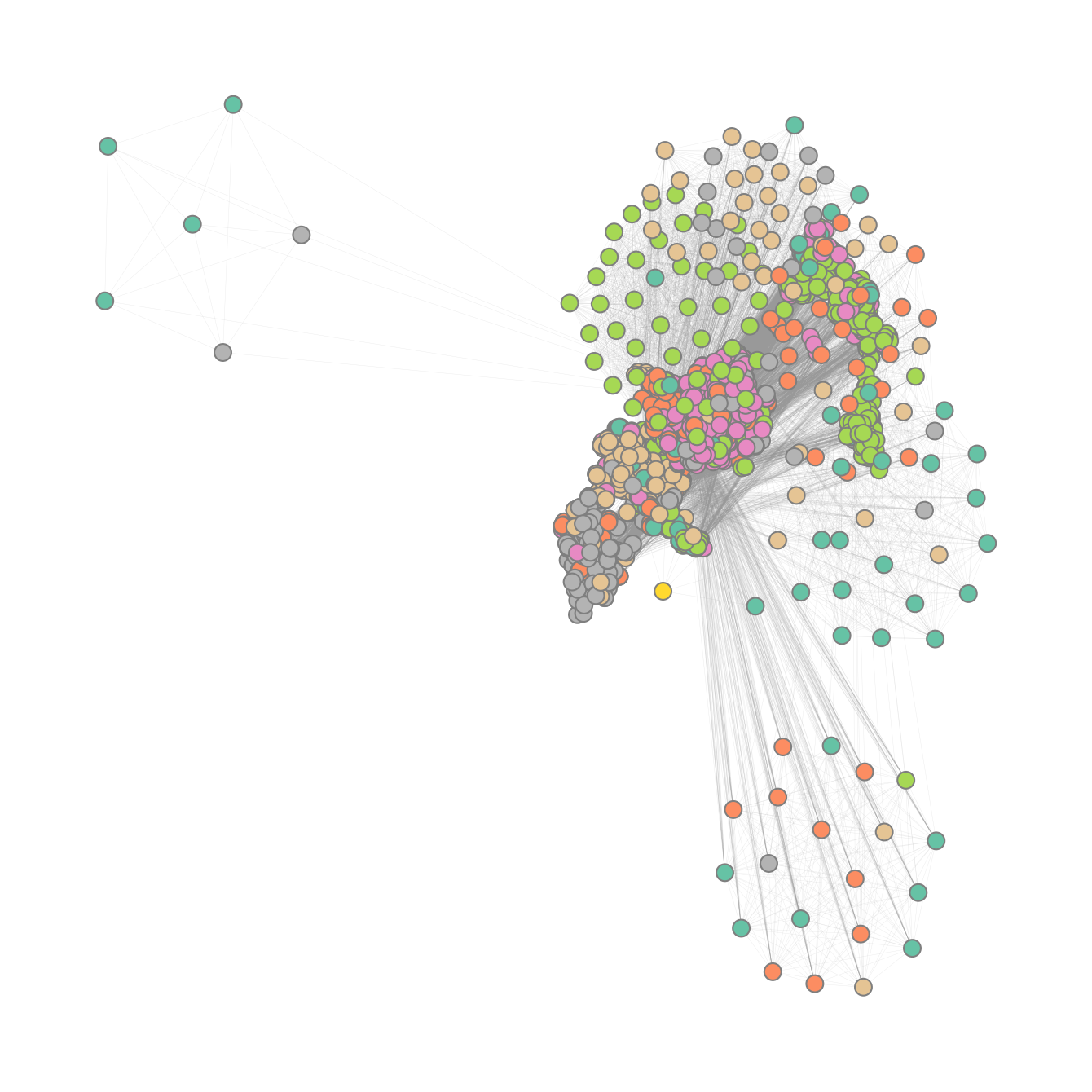}}
\\
    \subcaptionbox{\ds{squirrel} Raw Graph\label{fig:comm_squirrel_filtered}}%
    {\includegraphics[width=0.36\textwidth]{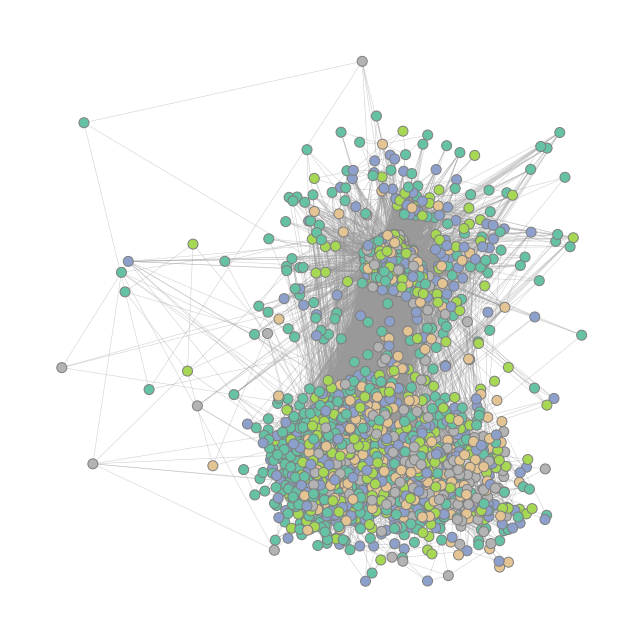}}
\hfil
    \subcaptionbox{\ds{squirrel} Label Graph\label{fig:colb_squirrel_filtered}}%
    {\includegraphics[width=0.36\textwidth]{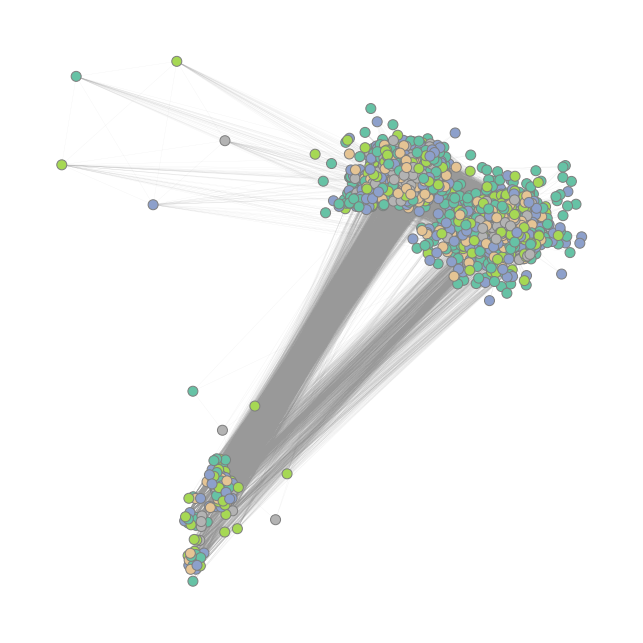}}
\caption{Visualization of the hierarchy of original and label graphs on realistic datasets. Color of each node denotes its class.}
\label{fig:comma}
% \vspace{-1.0em}
\end{figure*}

\iffalse
\begin{figure*}[tp]
\centering
    \subcaptionbox{\ds{cora} Raw\label{fig:comm_cora}}%
    {\includegraphics[width=0.162\textwidth]{figs/comm_cora.png}}
\hfil
    \subcaptionbox{\ds{cora} Label\label{fig:colb_cora}}%
    {\includegraphics[width=0.162\textwidth]{figs/colb_cora.png}}
\hfil
    \subcaptionbox{\ds{citeseer} Raw\label{fig:comm_citeseer}}%
    {\includegraphics[width=0.162\textwidth]{figs/comm_citeseer.png}}
\hfil
    \subcaptionbox{\ds{citeseer} Label\label{fig:colb_citeseer}}%
    {\includegraphics[width=0.162\textwidth]{figs/colb_citeseer.png}}
\hfil
    \subcaptionbox{\ds{squirrel} Raw\label{fig:comm_squirrel_filtered}}%
    {\includegraphics[width=0.162\textwidth]{figs/comm_squirrel_filtered.png}}
\hfil
    \subcaptionbox{\ds{squirrel} Label\label{fig:colb_squirrel_filtered}}%
    {\includegraphics[width=0.162\textwidth]{figs/colb_squirrel_filtered.png}}
\caption{Visualization of the hierarchy of original and label graphs on realistic datasets. Color of each node denotes its class.}
\label{fig:comma}
% \vspace{-1.0em}
\end{figure*}
\fi

\end{document}